\documentclass{article}
\usepackage{nips2004e,times,graphicx}
\def\half{{1\over 2}}
\def\A{{\cal A}}
\def\O{{\cal O}}

\def\u{\mbox{\bf u}}
\def\x{\mbox{\bf x}}
\def\y{\mbox{\bf y}}
\def\yy{\tilde{\mbox{\bf y}}}
\def\z{\mbox{\bf z}}

\def\c{{\bf c}}
\newcommand{\bsigma}{\mbox{\boldmath$\sigma$}}
\def\half{{1\over 2}}
\def\A{{\cal A}}
\def\O{{\cal O}}

\def\W{{\bf W}}

\title{Temporal-Difference Networks}

\author{
Richard S.~Sutton and Brian Tanner\\
Department of Computing Science\\
University of Alberta\\
Edmonton, Alberta, Canada T6G 2E8 \\
\texttt{\{sutton,btanner\}@cs.ualberta.ca}}

\begin{document}

\maketitle

\begin{abstract}
We introduce a generalization of temporal-difference (TD) learning to networks of interrelated predictions.
Rather than relating a single prediction to itself at a later time, as in conventional TD methods, a TD network relates each prediction in a set of predictions to other predictions in the set at a later time.  
TD networks can represent and apply TD learning to a much wider class of predictions than has previously been possible.  
Using a random-walk example, we show that these networks can be used to learn to predict by a fixed interval, which is not possible with conventional TD methods.  
Secondly, we show that if the inter-predictive relationships are made conditional on action, then the usual learning-efficiency advantage of TD methods over Monte Carlo (supervised learning) 
methods becomes particularly pronounced.  Thirdly, we demonstrate that TD networks can learn predictive state representations that enable exact solution of a non-Markov problem.  
A very broad range of inter-predictive temporal relationships can be expressed in these networks.  Overall we argue that TD networks represent a substantial extension of the abilities of TD methods and bring us closer to the goal of 
representing world knowledge in entirely predictive, grounded terms.  
\end{abstract}

\bigskip
Temporal-difference (TD) learning is widely used in reinforcement learning methods to learn moment-to-moment predictions of total future reward (value functions).  In this setting, TD learning is often simpler and more data-efficient than other methods.  But the idea of TD learning can be used more generally than it is in reinforcement learning.  TD learning is a general method for learning predictions whenever multiple predictions are made of the same event over time, value functions being just one example. The most pertinent of the more general uses of TD learning have been in learning models of an environment or task domain (Dayan, 1993; Kaelbling, 1993; Sutton, 1995; Sutton, Precup \& Singh, 1999).  In these works, TD learning is used to predict future values of many observations or state variables of a dynamical system.

The essential idea of TD learning can be described as ``learning a guess from a guess".  In all previous work, the two guesses involved were predictions of the same quantity at two points in time, for example, of the discounted future reward at successive time steps.  In this paper we explore a few of the possibilities that open up when the second guess is allowed to be different from the first.  

\goodbreak
To be more precise, we must make a distinction between the \emph{extensive definition\/} of a prediction, expressing its desired relationship to measurable data, and its \emph{TD definition}, expressing its desired relationship to other predictions.  In reinforcement learning, for example, state values are extensively defined as an expectation of the discounted sum of future rewards, while they are TD defined as the solution to the Bellman equation (a relationship to the expectation of the value of successor states, plus the immediate reward).  It's the same prediction, just defined or expressed in different ways.  In past work with TD methods, the TD relationship was always between predictions with identical or very similar extensive semantics.  In this paper we retain the TD idea of learning predictions based on others, but allow the predictions to have different extensive semantics.

\section{The Learning-to-predict Problem}

The problem we consider in this paper is a general one of learning to predict aspects of the interaction between a decision making agent and its environment.  
At each of a series of discrete time steps $t$, the environment generates an observation $o_t\in\O$, and the agent takes an action $a_t\in\A$.  
Whereas $\A$ is an arbitrary discrete set, we assume without loss of generality that $o_t$ can be represented as a vector of bits.  
The action and observation events occur in sequence, $o_1,a_1,o_2,a_2,o_3\cdots$, with each event of course dependent only on those preceding it.  
This sequence will be called \emph{experience}.  
We are interested in predicting not just each next observation but more general, action-conditional functions of future experience, as discussed in the next section.

In this paper we use a random-walk problem with seven states, with \textsl{left} and \textsl{right} actions available in every state:

\smallskip
\centerline {\includegraphics[width=5in]{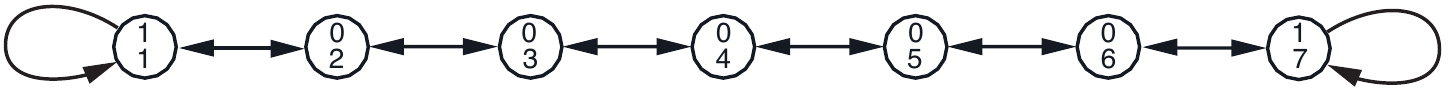}}

The observation upon arriving in a state consists of a special bit that is $1$ only at the two ends of the walk and, in the first two of our three experiments, seven additional bits explicitly indicating the state number (only one of them is 1).  This is a \emph{continuing} task:
reaching an end state does not end or interrupt experience.  Although the sequence depends deterministically on action, we assume that the actions are selected randomly with equal probability so that the overall system can be viewed as a Markov chain.

The TD networks introduced in this paper can represent a wide variety of predictions, far more than can be represented by a conventional TD predictor.  
In this paper we take just a few steps toward more general predictions.  In particular, we consider variations of the problem of prediction by a fixed interval.  This is one of the simplest cases that cannot otherwise be handled by TD methods.  For the seven-state random walk, we will predict the special observation bit some numbers of discrete steps in advance, first unconditionally and then conditioned on action sequences.

\bigskip
\goodbreak
\section{TD Networks}

A {\it TD network\/} is a network of nodes, each representing a single scalar prediction.  The nodes are interconnected by links representing the TD relationships among the predictions and to the observations and actions.   
These links determine the extensive semantics of each prediction---its desired or target relationship to the data.  They represent \emph{what} we seek to predict about the data as opposed to \emph{how} we try to predict it.  We think of these links as determining a set of \emph{questions\/} being asked about the data, and accordingly we call them the \emph{question network}.  A separate set of interconnections determines the actual computational process---the updating of the predictions at each node from their previous values and the current action and observation.  We think of this process as providing the \emph{answers\/} to the questions, and accordingly we call them the \emph{answer network}.  The question network provides targets for a learning process shaping the answer network and does not otherwise affect the behavior of the TD network.  It is natural to consider changing the question network, but in this paper we take it as fixed and given.

\begin{figure}[h]
\centerline {\includegraphics[width=5in]{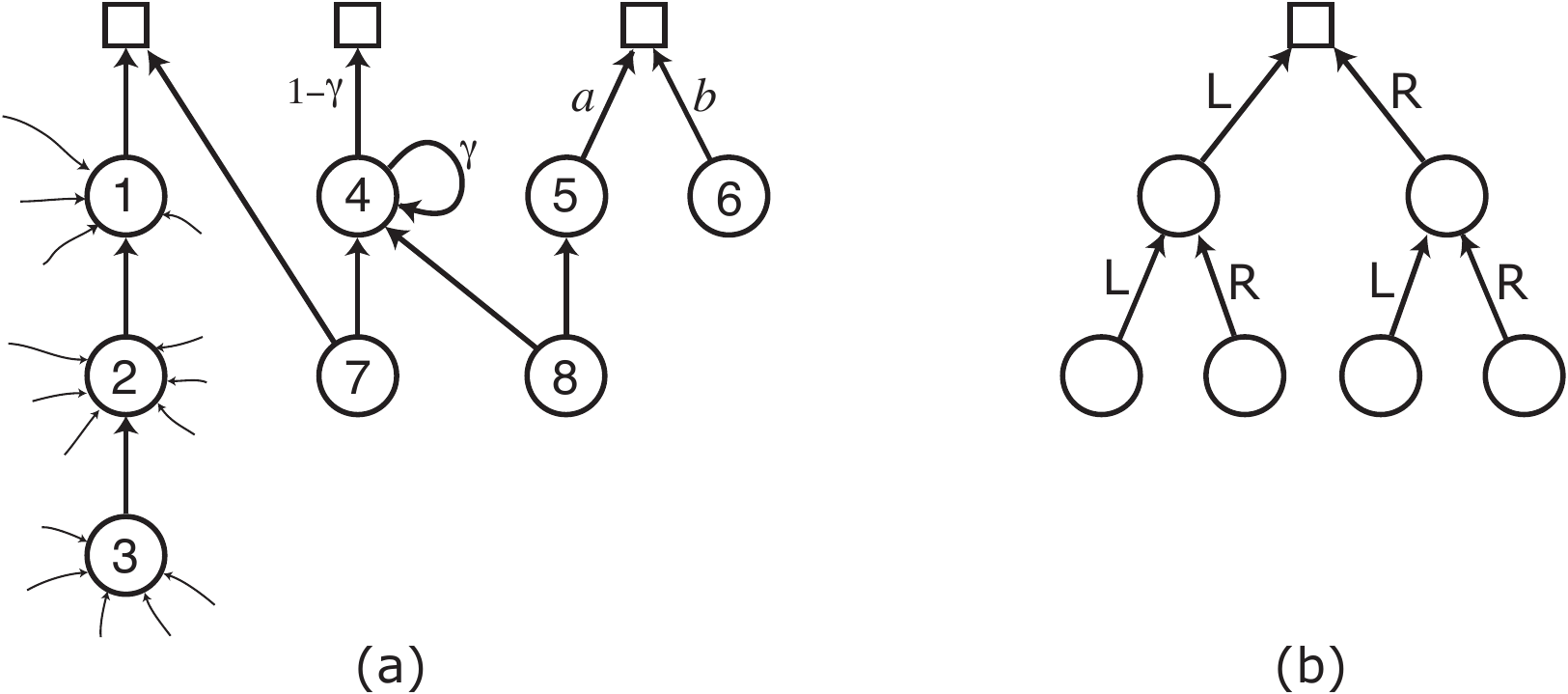}}
\caption{The question networks of two TD networks. (a) a question network discussed in the text, and (b) a depth-2 fully-action-conditional question network used in Experiments 2 and 3. Observation bits are represented as squares across the top while actual nodes of the TD network, corresponding each to a separate prediction, are below.  
The thick lines represent the question network and the thin lines in (a) suggest the answer network (the bulk of which is not shown). Note that all of these nodes, arrows, and numbers are completely different and separate from those representing the random-walk problem on the preceding page.}
\label{network}
\end{figure}

Figure \ref{network}a shows a suggestive example of a question network.  The three squares across the top represent three observation bits.  The node labeled 1 is directly connected to the first observation bit and represents a prediction that that bit will be 1 on the next time step.  The node labeled 2 is similarly a prediction of the expected value of node 1 on the next step.  Thus the extensive definition of Node 2's prediction is the probability that the first observation bit will be 1 two time steps from now.  Node 3 similarly predicts the first observation bit three time steps in the future. Node 4 is a conventional TD prediction, in this case of the future discounted sum of the second observation bit, with discount parameter $\gamma$.  Its target is the familiar TD target, the data bit {plus} the node's own prediction on the next time step (with weightings $1-\gamma$ and $\gamma$ respectively).  Nodes 5 and 6 predict the probability of the third observation bit being 1 \emph{if} particular actions \emph{a} or \emph{b} are taken respectively. Node 7 is a prediction of the average of the first observation bit and Node 4's prediction, both on the next step.  This is the first case where it is not easy to see or state the extensive semantics of the prediction in terms of the data.  Node 8 predicts another average, this time of nodes 4 and 5, and the question it asks is even harder to express extensively.  One could continue in this way, adding more and more nodes whose extensive definitions are difficult to express but which would nevertheless be completely defined as long as these local TD relationships are clear.  The thinner links shown entering some nodes are meant to be a suggestion of the entirely separate answer network determining the actual computation (as opposed to the goals) of the network.  In this paper we consider only simple question networks such as the left column of Figure \ref{network}a and of the action-conditional tree form shown in Figure \ref{network}b.

\bigskip

\newpage
More formally and generally, let $y^i_t\in[0,1]$, $i=1,\ldots,n$, denote the prediction of the $i$th node at time step $t$.  The column vector of predictions $\y_t=(y^1_t,\ldots,y^n_t)^T$ is updated according to a vector-valued function $\u$ with modifiable parameter $\W$:
  \begin{equation}
  \label{y}
  \y_{t}=\u(\y_{t-1},a_{t-1},o_{t},\W_{t}) \in \Re^n.
  \end{equation}
The update function $\u$ corresponds to the answer network, with $\W$ being the weights on its links. Before detailing that process, we turn to the question network, the defining TD relationships between nodes. The TD target $z^i_t$ for $y^i_t$ is an arbitrary function $z^i$ of the successive predictions and observations.  In vector form we have\footnote{In general, $\z$ is a function of all the future predictions and observations, but in this paper we treat only the one-step case.} 
  \begin{equation}
  \label{z}
  \z_t = \mbox{\bf z}(o_{t+1},\yy_{t+1}) \in \Re^n,
  \end{equation}
where $\yy_{t+1}$ is just like $\y_{t+1}$, as in (\ref{y}), except calculated with the \emph{old} weights before they are updated on the basis of $\z_t$:
  \begin{equation}
  \label{yy}
  \yy_{t}=\u(\y_{t-1},a_{t-1},o_{t},\W_{t-1}) \in \Re^n.
  \end{equation}
(This temporal subtlety also arises in conventional TD learning.)
For example, for the nodes in Figure \ref{network}a we have $z^1_t=o^1_{t+1}$, $z^2_t=y^1_{t+1}$, $z^3_t=y^2_{t+1}$, $z^4_t=(1-\gamma)o^2_{t+1} + \gamma y^4_{t+1}$, $z^5_t=z^6_t=o^3_{t+1}$, $z^7_t=\half o^1_{t+1} + \half y^4_{t+1}$, and $z^8_t=\half y^4_{t+1} + \half y^5_{t+1}$. The target functions $z^i$ are only part of specifying the question network.  The other part has to do with making them potentially conditional on action and observation.  For example, Node 5 in Figure \ref{network}a predicts what the third observation bit will be \emph{if} action \emph{a} is taken.
To arrange for such semantics we introduce a new vector $\c_t$ of \emph{conditions,\/} $c^i_t$, indicating the extent to which $y^i_t$ is held responsible for matching $z^i_t$, thus making the $i$th prediction conditional on $c^i_t$.  Each $c^i_t$ is determined as an arbitrary function $c^i$ of $a_t$ and $y_t$.  In vector form we have:
  \begin{equation}
  \label{bb}
  \c_t=\c(a_{t},\y_{t})\in[0,1]^n.
  \end{equation}
For example, for Node 5 in Figure \ref{network}a, $c^5_t=1$ if $a_t=a$, otherwise $c^5_t=0$.

Equations (\ref{z}--\ref{bb}) correspond to the \emph{question\/} network.
Let us now turn to defining $\u$, the update function for $\y_t$ mentioned earlier and which corresponds to the \emph{answer\/} network.  In general $\u$ is an arbitrary function approximator, but for concreteness we define it to be of a linear form
  \begin{equation}
  \label{ysigma}
  \y_{t}=\bsigma(\W_{t}\x_t)
  \end{equation}
where $\x_{t}\in\Re^m$ is a feature vector, $\W_t$ is an $n\times m$ matrix, and $\sigma$ is the $n$-vector form of the identity function (Experiments 1 and 2) or the S-shaped logistic function $\sigma(s)={1\over 1+e^{-s}}$ (Experiment 3).  The feature vector is an arbitrary function of the preceding action, observation, and node values:
  \begin{equation}
  \label{x}
  \x_{t} = \x(a_{t-1},o_{t},\y_{t-1}) \in \Re^m.
  \end{equation}
For example, $\x_t$ might have one component for each observation bit, one for each possible action (one of which is $1$, the rest $0$), and $n$ more for the previous node values $\y_{t-1}$.
The learning algorithm for each component $w^{ij}_t$ of $\W_t$ is
  \begin{equation}
  \label{eq:learn}
   w^{ij}_{t+1} - w^{ij}_{t} = \alpha (z^i_t - y^i_t)c^i_t {\partial y^i_t\over\partial w^{ij}_{t}},  
  \end{equation}
where $\alpha$ is a step-size parameter.  The timing details may be clarified by writing the sequence of quantities in the order in which they are computed:
  \begin{equation}
  \label{timing}
%  \y_{t-1} \: a_{t-1} \: \c_{t-1} \: o_t \: \x_t \: \yy_{t} \: \z_{t-1} \: \W_{t} \: 
\y_{t} \: a_{t} \: \c_t \: o_{t+1} \: \x_{t+1} \: \yy_{t+1} \: \z_t \: \W_{t+1} \: \y_{t+1}.
  \end{equation}
Finally, the target in the extensive sense for $\y_t$ is
  \begin{equation}
  \label{eq:y*}
  \y^*_{t}=E_{t,\pi}\left\{(1-\c_{t})\cdot\y^*_{t} + \c_{t}\cdot\mbox{\bf z}(o_{t+1},\y^*_{t+1})\right\},   
  \end{equation}
where $\cdot$ represents component-wise multiplication and $\pi$ is the policy being followed, which is assumed fixed.
 
\section{Experiment 1: $n$-step Unconditional Prediction}

In this experiment we sought to predict the observation bit precisely $n$ steps in advance, for $n=1$, 2, 5, 10, and 25.  In order to predict $n$ steps in advance, of course, we also have to predict $n-1$ steps in advance, $n-2$ steps in advance, etc., 
all the way down to predicting one step ahead.  This is specified by a TD network consisting of a single chain of predictions like the left column of Figure \ref{network}a, but of length 25 rather than 3.  
Random-walk sequences were constructed by starting at the center state and then taking random actions for 50, 100, 150, and 200 steps (100 sequences each).

We applied a TD network and a corresponding Monte Carlo method to this data.  The Monte Carlo method learned the same predictions, but learned them by comparing them to the actual outcomes in the sequence (instead of $z^i_t$ in (\ref{eq:learn})).  This involved significant additional complexity to store the predictions until their corresponding targets were available.  Both algorithms used feature vectors of 7 binary components, one for each of the seven states, all of which were zero except for the one corresponding to the current state.  Both algorithms formed their predictions linearly ($\sigma(\cdot)$ was the identity) and unconditionally ($c^i_t=1\,\,\forall i,t$).

In an initial set of experiments, both algorithms were applied online with a variety of values for their step-size parameter $\alpha$.  Under these conditions we did not find that either algorithm was clearly better in terms of the mean square error in their predictions over the data sets.  We found a clearer result when both algorithms were trained using batch updating, in which weight changes are collected ``on the side" over an experience sequence and then made all at once at the end, and the whole process is repeated until convergence.  Under batch updating, convergence is to the same predictions regardless of initial conditions or $\alpha$ value (as long as $\alpha$ is sufficiently small), which greatly simplifies comparison of algorithms.  The predictions learned under batch updating are also the same as would be computed by least squares algorithms such as LSTD($\lambda$) (Bradtke \& Barto, 1996; Boyan, 2000; Lagoudakis \& Parr, 2003). The errors in the final predictions are shown in Table 1.

For 1-step predictions, the Monte-Carlo and TD methods performed identically of course, but for longer predictions a significant difference was observed.  The RMSE of the Monte Carlo method increased with prediction length whereas for the TD network it decreased.  The largest standard error in any of the numbers shown in the table is 0.008, so almost all of the differences are statistically significant.  TD methods appear to have a significant data-efficiency advantage over non-TD methods in this prediction-by-$n$ context (and this task) just as they do in conventional multi-step prediction (Sutton, 1988).

\begin{table}[h]
\begin{center}{
\begin{tabular}{|c||c||c|c||c|c||c|c||c|c|}
\hline
\multicolumn{1}{|c||}{Time} & \multicolumn{1}{c||}{1-step} &
\multicolumn{2}{c||}{2-step}  &  \multicolumn{2}{c||}{5-step} &
\multicolumn{2}{c||}{10-step} &  \multicolumn{2}{c|}{25-step} \\
%\cline{2-10}
 Steps & MC/TD & MC & TD & MC & TD & MC & TD & MC & TD \\
\hline
50 & 0.205 & 0.219 & 0.172 & 0.234 & 0.159 & 0.249 & 0.139 & 0.297
& 0.129 \\
100 &0.124 & 0.133 & 0.100 & 0.160 & 0.098 & 0.168 & 0.079 &
0.187 & 0.068 \\
150 &0.089 & 0.103 & 0.073 & 0.121 & 0.076 & 0.130 & 0.063 &
0.153 & 0.054 \\
200 &0.076 & 0.084 & 0.060 & 0.109 & 0.065 & 0.112 & 0.056 &
0.118 & 0.049 \\
\hline
\end{tabular}
}\end{center}
\vspace{-.05in}
\caption{RMSE of Monte-Carlo and TD-network predictions of various lengths and for increasing amounts of training data on the random-walk example with batch updating. }
\label{xp1}
 \end{table}

\section{Experiment 2: Action-conditional Prediction}

The advantage of TD methods should be greater for predictions that apply only when the experience sequence unfolds in a particular way, such as when a particular sequence of actions are made.  In a second experiment we sought to 
learn $n$-step-ahead predictions conditional on action selections.  The question network for learning all 2-step-ahead predictions is shown in Figure \ref{network}b.  The upper two nodes predict the observation bit conditional on taking a left action (L) or a right action (R).  The lower four nodes correspond to the two-step predictions, e.g., the second lower node is the prediction of what the observation bit will be if an L action is taken followed by an R action.  These predictions are the same as the \emph{e-tests} used in some of the work on predictive state representations (Littman, Sutton \& Singh, 2002; Rudary \& Singh, 2003).

In this experiment we used a question network like that in Figure~\ref{network}b except of depth four, consisting of 30 (2+4+8+16) nodes.  The conditions for each node were set to $0$ or $1$ depending on whether the action taken on the step matched that indicated in the figure.
The feature vectors were as in the previous experiment.  Now that we are conditioning on action, the problem is deterministic and $\alpha$ can be set uniformly to $1$.  A Monte Carlo prediction can be learned only when its corresponding action sequence occurs in its entirety, but then it is complete and accurate in one step.
The TD network, on the other hand, can learn from incomplete sequences but must propagate them back one level at a time.  First the one-step predictions must be learned, then the two-step predictions from them, and so on.
The results for online and batch training are shown in Tables \ref{xp2online} and \ref{xp2batch}.

As anticipated, the TD network learns much faster than Monte Carlo with both online and batch updating.  Because the TD network learns its $n$ step predictions based on its $n-1$ step predictions, it has a clear advantage for this task.  Once the TD Network has seen each action in each state, it can quickly learn any prediction 2, 10, or 1000 steps in the future.  Monte Carlo, on the other hand, must sample actual sequences, so each exact action sequence must be observed.  

\begin{table}[h]
\begin{center}
\begin{tabular}{|c||c||c|c||c|c||c|c|}
\hline
 \multicolumn{1}{|c||}{ } & \multicolumn{1}{c||}{1-Step} & \multicolumn{2}{c||}{2-Step}  &  \multicolumn{2}{c||}{3-Step} & \multicolumn{2}{c|}{4-Step}\\
% \cline{2-8}
 Time Step & MC/TD & MC & TD & MC & TD & MC & TD\\
\hline
100 & 0.153 & 0.222 & 0.182 & 0.253 & 0.195 & 0.285 & 0.185\\
200 & 0.019 & 0.092 & 0.044 & 0.142 & 0.054 & 0.196 & 0.062\\
300 & 0.000 & 0.040 & 0.000 & 0.089 & 0.013 & 0.139 & 0.017\\
400 & 0.000 & 0.019 & 0.000 & 0.055 & 0.000 & 0.093 & 0.000\\
500 & 0.000 & 0.019 & 0.000 & 0.038 & 0.000 & 0.062 & 0.000\\
\hline
\end{tabular}
\end{center}
\vspace{-.05in}
\caption{RMSE of the action-conditional predictions of various lengths for Monte-Carlo and TD-network methods on the random-walk problem with online updating.}
\label{xp2online}
\end{table}

\begin{table}[h]
\begin{center}
 \begin{tabular}{|c|c|r|}
 \hline
 Time Steps & MC & TD~~~~ \\
 \hline
 50 &53.48\% & 17.21\% \\
 100 & 30.81\% & 4.50\% \\
 150 & 19.26\% & 1.57\% \\
 200 & 11.69\% & 0.14\% \\
 \hline
 \end{tabular}
\end{center}
\vspace{-.05in}
\caption{Average proportion of incorrect action-conditional predictions for batch-updating versions of Monte-Carlo and TD-network methods, for various amounts of data, on the random-walk task.  All differences are statistically significant.}
\label{xp2batch}
\end{table}

\section{Experiment 3: Learning a Predictive State Representation}

Experiments 1 and 2 showed advantages for TD learning methods in Markov problems.  The feature vectors in both experiments provided complete information about the nominal state of the random walk.  In Experiment 3, on the other hand, we applied TD networks to a non-Markov version of the random-walk example, in particular, in which only the special observation bit was visible and not the state number.  In this case it is not possible to make accurate predictions based solely on the current action and observation; the previous time step's predictions must be used as well.  

As in the previous experiment, we sought to learn $n$-step predictions using action-conditional question networks of depths 2, 3, and 4.  The feature vector $\x_t$ consisted of three parts:
a constant 1, four binary features to represent the \emph{pair} of action $a_{t-1}$ and observation bit $o_t$, and $n$ more features corresponding to the components of $\y_{t-1}$.  The features vectors were thus of length $m=11, 19, \mbox{and~}\, 35$ for the three depths. In this experiment, $\sigma(\cdot)$ was the S-shaped logistic function.  The initial weights $\W_{0}$ and predictions $\y_{0}$ were both $0$.  

\begin{figure}[h]
\centerline {\includegraphics[width=5in]{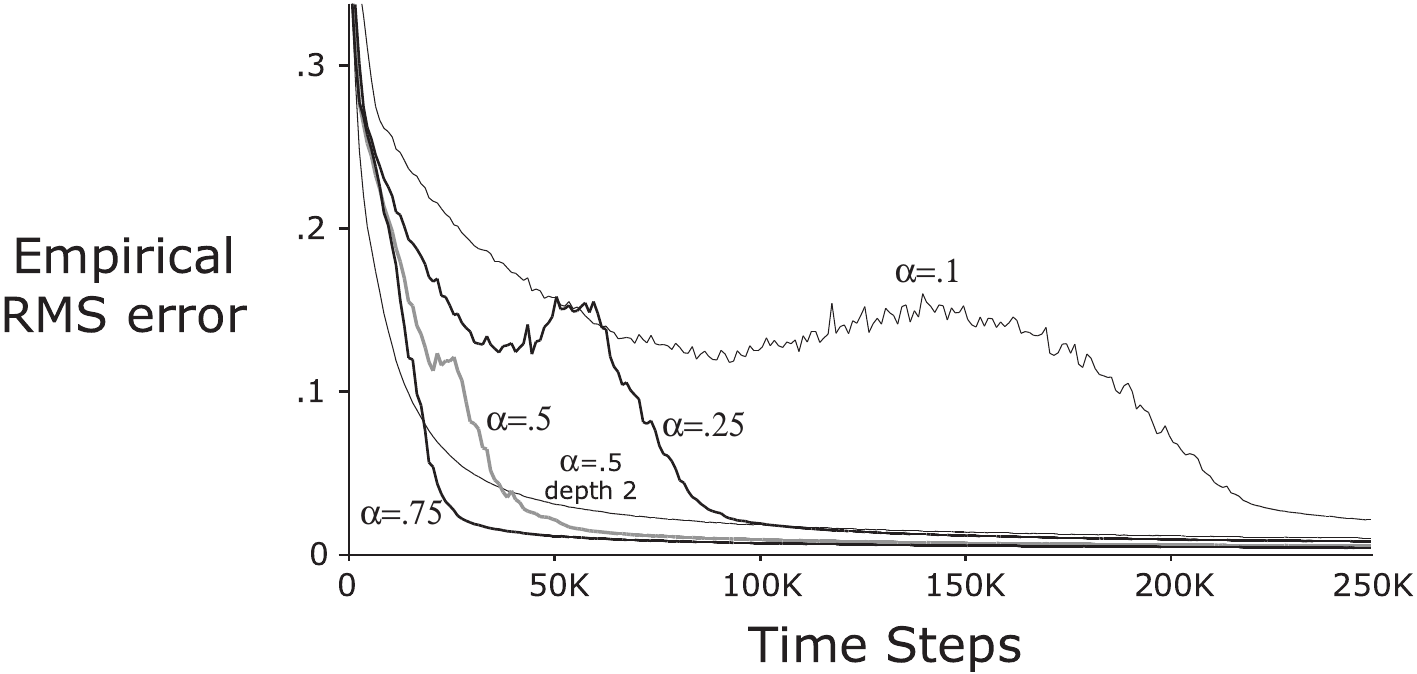}}
\vspace{-.1in}
\caption{Prediction performance on the non-Markov random walk with depth-4 TD networks (and one depth-2 network) with various step-size parameters, 
averaged over 50 runs and 1000 time-step bins. The ``bump" most clearly seen with small step sizes is reliably present and may be due to predictions of different lengths being learned at different times.
}
\label{exp3}
\end{figure}

Fifty random-walk sequences were constructed, each of 250,000 time steps, and presented to TD networks of the three depths, with a range of step-size parameters $\alpha$.  We measured the RMSE of all predictions made by the networks (computed from knowledge of the task) and also the ``empirical RMSE," the error in the one-step prediction for the action actually taken on each step.  We found that in all cases the errors approached zero over time, showing that the problem was completely solved.
Figure \ref{exp3} shows some representative learning curves for the depth-2 and depth-4 TD networks.

In ongoing experiments on other non-Markov problems we have found that TD networks do not always find such complete solutions.  Other problems seem to require more than one step of history information (the one-step-preceding action and observation), though less than would be required using history information alone.  Our results as a whole suggest that TD networks may provide an effective alternative learning algorithm for predictive state representations (Littman et al., 2000).  Previous algorithms have been found to be effective on some tasks but not on others (e.g, Singh et al., 2003; Rudary \& Singh, 2004; James \& Singh, 2004).  More work is needed to assess the range of effectiveness and learning rate of TD methods vis-a-vis previous methods, and to explore their combination with history information.

\section{Conclusion}

TD networks suggest a large set of possibilities for learning to predict, and in this paper we have begun exploring the first few.  Our results show that even in a fully observable setting there may be significant advantages to TD methods when learning TD-defined predictions.  Our action-conditional results show that TD methods can learn dramatically faster than other methods.  TD networks allow the expression of many new kinds of predictions whose extensive semantics is not immediately clear, but which are ultimately fully grounded in data.  It may be fruitful to further explore the expressive potential of TD-defined predictions.

Although most of our experiments have concerned the representational expressiveness and efficiency of TD-defined predictions, it is also natural to consider using them as state, as in predictive state representations.  Our experiments suggest that this is a promising direction and that TD learning algorithms may have advantages over previous learning methods. Finally, we note that adding nodes to a question network produces new predictions and thus may be a way to address the discovery problem for predictive representations.

\subsubsection*{Acknowledgments}

The authors gratefully acknowledge the ideas and encouragement they have received in this work from Satinder Singh, Doina Precup, Michael Littman, Mark Ring, Vadim Bulitko, Eddie Rafols, Anna Koop, Tao Wang, and all the members of the rlai.net group.

\subsubsection*{References}

{\small
Boyan, J.\ A.  (2000). Technical update: Least-squares temporal difference learning. \textit{Machine Learning 49}:233--246.

Bradtke, S.\ J.\ and Barto, A.\ G. (1996). Linear least-squares algorithms for temporal difference learning.  \textit{Machine Learning
22}(1/2/3):33--57.

Dayan, P. (1993). Improving generalization for temporal difference learning: The successor representation. \emph{Neural Computation 5}(4):613--624.

James, M.\ and Singh, S. (2004). Learning and discovery of predictive state representations in dynamical systems with reset. In {\em Proceedings of the Twenty-First International Conference on Machine Learning}, pages 417--424.

Kaelbling, L.~P. (1993).
\newblock Hierarchical learning in stochastic domains: {P}reliminary results.
\newblock In {\em Proceedings of the Tenth International Conference on Machine
  Learning}, pp.~167--173.

Lagoudakis, M.~G.\ and Parr, R. (2003). Least-squares policy iteration.
\emph{Journal of Machine Learning Research 4}(Dec):1107--1149.

Littman, M.\ L., Sutton, R.\ S.\ and Singh, S.  (2002). Predictive representations of state.  In \textit{Advances In Neural Information Processing Systems 14}:1555--1561.

Rudary, M.\ R.\ and Singh, S.  (2004). A nonlinear predictive state representation.
In \textit{Advances in Neural Information Processing Systems 16}:855--862.

Singh, S., Littman, M.\ L., Jong, N.\ K., Pardoe, D.\ and Stone, P. (2003) Learning predictive state  representations.
In \textit{Proceedings of the Twentieth Int.\ Conference on Machine Learning}, pp.~712--719.

Sutton, R.~S. (1988).
\newblock Learning to predict by the methods of temporal differences.
\newblock {\em Machine Learning 3}:9--44.

Sutton, R.~S. (1995).
\newblock {TD} models: {M}odeling the world at a mixture of time scales.
\newblock In A.~Prieditis and S.~Russell (eds.), {\em Proceedings of the
  Twelfth International Conference on Machine Learning}, pp.~531--539.
Morgan Kaufmann,
  San Francisco.
  
Sutton, R.~S., Precup, D. and Singh, S. (1999).
\newblock Between {MDP}s and semi-{MDP}s: {A} framework for temporal
  abstraction in reinforcement learning.
\newblock {\em Artificial Intelligence 112}:181--121.

}

\end{document}